\documentclass{article}

\usepackage{arxiv}

\usepackage[utf8]{inputenc} % allow utf-8 input
\usepackage[T1]{fontenc}    % use 8-bit T1 fonts
\usepackage{hyperref}       % hyperlinks
\usepackage{url}            % simple URL typesetting
\usepackage{booktabs}       % professional-quality tables
\usepackage{amsfonts}       % blackboard math symbols
\usepackage{nicefrac}       % compact symbols for 1/2, etc.
\usepackage{microtype}      % microtypography
\usepackage{lipsum}		% Can be removed after putting your text content
\usepackage{graphicx}
\usepackage{natbib}
\usepackage{doi}

\usepackage{ragged2e}
\usepackage{amsmath}
\usepackage{pythontex}
\usepackage{soul}

\graphicspath{ {./images/} }

\title{Masterful: A training platform for computer vision models}

\date{April 25th, 2022}	% Here you can change the date presented in the paper title
\author{ \href{https://scholar.google.com/citations?user=Dz4sfKYAAAAJ}{Samuel Wookey}\thanks{Affiliated with Masterful AI at time of relevant work. Corresponding author: yaoshiang@masterfulai.com.}
	\And
	\href{https://orcid.org/0000-0003-0822-6269}{Yaoshiang Ho} \footnotemark[1]
	\And
	\href{https://scholar.google.com/citations?user=EaceQz0AAAAJ}{Tom Rikert} \footnotemark[1]
	\And
	Juan David Gil Lopez  \footnotemark[1]
	\And
	Juan Manuel Muñoz Betancur\footnotemark[1]	
	\And
	Santiago Cortes \footnotemark[1]	
	\And
	Ray Tawil \footnotemark[1]	
	\And
	Aaron Sabin \footnotemark[1]
	\And
	\href{https://scholar.google.com/citations?user=jLZn3dYAAAAJ}{Jack Lynch}  \footnotemark[1]
	\And
	Travis Harper \footnotemark[1]
	\And
	\href{https://scholar.google.com/citations?user=A8ILPysAAAAJ}{Nikhil Gajendrakumar} \footnotemark[1]
}

\date{}

\renewcommand{\headeright}{ }
\renewcommand{\undertitle}{ }
\renewcommand{\shorttitle}{ }

\begin{document}
\maketitle

\begin{abstract}
\textbf{Masterful} is a software platform to train deep learning computer vision models. Data and model architecture are inputs to the platform, and the output is a trained model. The platform's primary goal is to maximize a trained model's accuracy, which it achieves through its regularization and semi-supervised learning implementations. The platform's secondary goal is to minimize the amount of manual experimentation typically required to tune training hyperparameters, which it achieves via multiple metalearning algorithms which are custom built to control the platform's regularization and semi-supervised learning implementations. The platform's tertiary goal is to minimize the computing resources required to train a model, which it achieves via another set of metalearning algorithms which are purpose built to control Tensorflow's optimization implementations. The platform builds on top of Tensorflow's data management, architecture, automatic differentiation, and optimization implementations.
\end{abstract}

% keywords can be removed

\section{Introduction}
The field of computer vision (CV) has relied on deep learning to achieve state-of-the-art results since the breakthrough performance of AlexNet \citep{krizhevsky2012imagenet}. Numerous research advancements, primarily in model architecture but also in training techniques, have continued to improve CV model performance. However, practitioners attempting to apply research often run into challenges:

\begin{itemize}
	\item Unique attributes of research datasets often do not apply to real world datasets.  For example, CIFAR-10 and Imagenet are balanced,  have high entropy,  and high cardinalities \citep{krizhevsky2009learning, deng2009imagenet}. Research built on these datasets sometimes need adjustments to deal with the class imbalanced, low entropy, and low cardinality datasets common in practice. 
	\item An emphasis on model architectures leaves training techniques a less studied field. For example, while data augmentation was included in AlexNet, the arguably first significant paper to study data augmentation policies in depth occurred in 2019, 7 years after the publication of AlexNet \citep{cubuk2018autoaugment}. And the interplay of adaptive optimizers and weight decay, a regularization technique used in the majority of state of the art CV research papers, was misunderstood until 2019 \citep{loshchilov2017decoupled}. 
	\item The training of models still requires extensive hyperparameter tuning. While some automatic hyperparameter tuning techniques exist, the major approach is still manually guessing-and-checking hyperparameters \citep{karpathy}.
	\item Deploying models into inference adds additional model size and latency constraints. For example, when deploying deep learning CV models on to mobile devices, it is common to take a well trained model and reduce its size via pruning and fixed point quantization \citep{goyal2021fixed}.

\end{itemize}

The Masterful platform seeks to solve these problems to enable practitioners to more easily apply deep learning to their CV problems. Masterful is guided by the following design choices.

\begin{itemize}
	\item A focus on a domain of problems that encompasses the vast majority of computer vision tasks in practice, but is nevertheless a small minority of the full expressiveness of underlying vectorized computing, neural network design, and back propagation based ML platforms like Tensorflow and PyTorch \citep{tensorflow2015-whitepaper}. 
	\item A primary goal of model accuracy (or other goodness-of-fit measures) by implementing state-of-the-art regularization and semi-supervised learning (SSL) algorithms. We believe that this goal requires building an integrated platform rather than implementing a collection of algorithms in isolation because algorithms are not additive. Like an athlete who overtrains to the point of injury, applying two such algorithms may be worse than no algorithms at all \citep{muller2019does}.
	\item A secondary goal of reducing developer time spent on hyper-parameter tuning. Key hyperparameters of research papers are often grid-searched or manually guess-and-checked behind the scenes. And when applying a technique to a new dataset or model architecture, new hyperparameters often must be discovered. Masterful's goal is to minimize the time spent on hyperparameter tuning by developing purpose-built algorithms to discover these hyperparameters. These algorithms range from highly optimized implementations of search algorithms specific to small subsets of the hyperparameters, to analytical, closed-form solutions. 
	\item A tertiary goal of minimizing compute resources, e.g. GPU-hours. Compute resources are a major source of training cost, our experiences has been that most practitioners utilize less than half of their compute resources. Common mistakes we have seen in practice include using antiquated optimizers, confusing regularization and optimization, using hyperparameters like Batch Normalization momentum inappropriate for their datasets, inappropriately low batch sizes, and inappropriately low learning rates. We believe that the goal of optimizers like Adam or LAMB have the sole purpose to minimize compute resources by converging a model's weights to the minimum loss value on training data in the minimum amount of time. It is outside the scope of optimization to focus on generalization (e.g. accuracy on test data), which is the domain of regularization. This can be confusing in practice - for example many optimizers implement decoupled weight decay, a regularization technique. In our view, this simply means that it was convenient to implement a specific type of regularization and optimization in the same software package. Indeed, decoupled weight decay has also been implemented as a separate module, divorced from the optimizer \citep{loshchilov2017decoupled}. Optimizers are the only way to find a minimum loss value but they are not the only approach to speeding up the overall task of optimization. For example early stopping is also a method to reduce the use of compute resources \citep{prechelt1998early}. 

\end{itemize}

Masterful is currently built to support model architectures implemented as Tensorflow 2 and Tensorflow Object Detection API formats. And data formats implemented as Tensorflow Tensors, \texttt{tf.data.Dataset}, and Numpy arrays \citep{tensorflow2015-whitepaper}. None of our algorithms are format specific. We plan to extend to PyTorch formats in the near future. 

Masterful supports these common CV use cases:

\begin{itemize}
	\item Binary classification
	\item Single-label classification (e.g. ImageNet, CIFAR-10)
	\item Multi-label classification
	\item Detection (e.g. Pascal VOC, MS-COCO)
	\item Keypoint Detection (coming soon)
	\item Semantic Segmentation
	\item Instance Segmentation (coming soon)
\end{itemize}
	
\section{Prior Work}

\citet{bello2021revisiting} (ResNet-RS) set a framework that improvements in model accuracy could be grouped into training techniques (what we call optimization), regularization, and model architecture. It also proposed the view that model architecture was given undue credit in prior work when in fact regularization and training techniques were responsible for at least some of the apparently improvement from a model architecture. 

\citet{loshchilov2017decoupled} (Decoupled Weight Decay). First, it teased out the difference between regularization and optimization, despite the fact that both can be implemented in a software package known as an "optimizer", such as Adam \citep{kingma2014adam}. Second, it resolved a long-standing mystery as to why the Adam optimizer did not produce SOTA results. Finally, its conclusions pointed to training techniques in general as an understudied field, since many SOTA advancements in model architecture including AlexNet, GoogLeNet, Resnet, and EfficientNet did include weight decay, but without the same in-depth study of the subject. 

\citet{cui2019class} gave us an early hint that information theoretic approaches would be necessary to understand the amount of information in training data.

AutoAugment \citet{cubuk2019autoaugment} and RandAug \citet{cubuk2020randaugment} were major, significant studies of data augmentation, a subset of regularization. Both attempted to search the high dimensional space of data augmentations. RandAug in particular attempted to collapse this space into a tractable size, on the order of 100 full training runs rather than the 15,000 of AutoAug. Our work with RandAug and its shortcoming in datasets other than ImageNet led us to develop our own search algorithms to further reduce this search space while better adapting it to arbitrary model architectures and datasets. 

Cutmix, Mixup, Label Smoothing Regularization, and Distillation all used soft labels, which perhaps contain more knowledge (aka "Dark Knowledge") than hard labels \citep{yun2019cutmix, zhang2017mixup,  szegedy2016rethinking, hinton2015distilling, muller2019does}. 

Frechet Inception Distance introduced a powerful semantic similarity metric, a simplified variation of which we have implemented to help collapse the search space of data augmentations by orders of magnitude compared to AutoAug \citep{heusel2017gans}. 

\citet{mccandlish2018empirical} (Gradient Noise Scale) showed that statistical analysis of gradients wasn't limited solely to the field of optimizers, but also towards an information theoretic approach to determining the optimal batch size, rather than a heuristical approach. This also leads to accurate sizing of compute resources. 

\citet{smith2017cyclical} showed an empirical estimate of a max learning rate. 

Self-Distillation as Instance-Specific Label Smoothing proposed the view that soft labels are actually a form of Maximum A Posteriori optimization \citep{zhang2019your}.

\citet{ho2019real} describes the frequentist philosophy that underlies deep learning via Maximum Likelihood Estimation. 

Snorkel's description of weak labeling combined with Distillation led us to our view that so-called "supervised training" is nothing more than distillation, where a target model is distilling the knowledge found in a single or an ensemble of larger, more powerful models: humans \citep{ratner2017snorkel}.

Noisy Student Training \citet{xie2020self}, SimCLR \citet{chen2020simple}, and Barlow Twins \citet{zbontar2021barlow} all established simple approaches to set SOTA results in Semi-Supervised Learning. 

\citet{gidaris2018unsupervised} provided an easy to reproduce example of unsupervised pretextual pretraining, the same framework as SimCLR, Barlow Twins, and other contrastive approaches.

\section{Model Architecture}

The Masterful platform takes as input a model architecture. This allows the user the control a specific architecture that will be appropriate for their inference constraints (size, latency, cost). 

In general, it's our belief that a modern family of architectures like ResNet-RS or EfficientNet will be nearly optimal for the vast majority of problems. In future work, we plan to simplify the task of model architecture selection through rules based on power laws. 

We are encouraged by work in transformers pretrained through generative techniques, which may help reduce the translation invariance assumption / inductive bias, which may in turn prove to be superior backbones for detection and segmentation than traditional convolutional neural networks \citep{carion2020end, bao2021beit}. 

\section{Data}

The Masterful platform takes data as an input. Since a key implementation of Masterful is SSL, we define data not only as labeled data (images and labels), but also unlabeled data (raw images without labels). 

Transfer learning can also be viewed as an approach to extract information from multiple sources of data \citep{pratt1992discriminability}. For example, by pretraining a backbone using a task like supervised pretraining or an unsupervised / self-supervised technique like SimCLR using Imagenet, then fine-tuning with labeled data from the target domain, we are essentially training a model using two sources of data: Imagenet and the target domain. 

\section{Meta-learning}

Masterful contains multiple meta-learners to discover the hyper-parameters  to control Regularization, SSL, and Optimization. We do not use any black box optimizer, such as those based on Bayesian optimization, because we believe that each hyper-parameter is best discovered using it's own meta-learning algorithm. Our meta-learners include closed form solutions, grid search on pretextual problems, beam search, greedy search, and reinforcement learning. 

We include discussion of our specific meta-learning algorithms in the respective sections of Regularization, SSL, and Optimization. 

\section{Regularization}

Masterful contains multiple algorithms to regularize a model. We define regularization as achieving accuracy (or other goodness-of-fit measures) on data the model has never seen, which the test dataset represents. This typically means preventing the model from overfitting to the training dataset.

The regularization methods included in Masterful include:

\begin{itemize}
	\item Data Augmentation, or image transforms. One view of data augmentation is that it reproduces invariances found in the image capture process. For example, zooming in and out of an image is a common data augmentation. If an object is a dog from five meters away from a camera, then that object is still a dog if the camera is ten meters away. Yet each camera captures a different image: the closer camera captures an image with a larger dog. This data augmentation technique obviously fails for abnormal ranges, such as microscopic ranges.
	\item Soft label techniques. Focusing on the labels rather than the images, forcing labels into continuous distributions like $(0.6, 0.4)$ appear to help regularize models as well in some cases.
	\item Weight Decay focuses on the weights of a models itself, based on the basic assumption from Rumelhart: "...the simplest most robust network which accounts/or a data set will, on average, lead to the best generalization to the population from which the training set has been drawn" \citep{hanson1988comparing}. Weight decay is generally a single parameter, and yet modern optimizers recognize that different weights or layers of a model benefit from differential learning rates. To meta-learn a decay rate per weight or layer would be intractable, but Masterful includes a novel technique to decay weights dynamically and separately from back-prop. 
	\item Dropout is a commonly used regularization technique that in one interpretation by it's original authors attempts to efficiently approximate a "Bayesian gold standard" \citep{JMLR:v15:srivastava14a}. 
\end{itemize}

Regularization inside the Masterful platform involves picking the right techniques with the right magnitude. Given  transforms $T$ and magnitudes $M$, the search space is of size $|T|^{|M|}$. In practice this can easily grow to a search space of $10^{10}$ values, which is obviously too large of a search space to tractably search via brute force. Therefore the central challenge is not any individual technique, but finding the right hyperparameters among all possible values. 

Transforms are not independent of each other, so a greedy approach of finding an optimal magnitude for an individual transform, next stacking on the next transform, etc., despite having a search space of only $ |T| \times |M| $, is ineffective. 

In order to reduce the search space, we analyze each tuple $(t,m)$ where $ t \in T$ and $m \in M$ using a simplified variation of Frechet Inception Distance and group each tuple by its distance \citep{heusel2017gans} into groups $G$. In the extreme, this grouping would allow us to reduce the search space to simply $|G|$. 

The implementations of the transforms are designed to run on GPU, making them also serve the goal of optimization. 

\section{Semi-Supervised Learning or SSL}

SSL is the ability for a CV model to learn from both labeled and unlabeled images.

Masterful trains your model using unlabeled data through SSL techniques. Masterful’s approach draws from two of the three major lineages of SSL: feature reconstruction and contrastive learning of representations. (Masterful does not currently include techniques from the third major lineage, generative techniques aka image reconstruction). Three state-of-the-art papers broadly define the techniques included in Masterful: Noisy Student Training, SimCLR, and Barlow Twins \citep{xie2020self, chen2020simple, zbontar2021barlow}.

The central challenge of productizing SSL is that research problems are often defined narrowly and therefore research results are not robust to messy, real world conditions. Defining a narrow problem like “classification on Imagenet on Resnet50” allows many hyperparameters to be asserted. Masterful generalizes the basic concepts from SOTA research to additional tasks like detection and segmentation, arbitrary data domains like overhead geospatial, and additional model types.

\section{Optimization}

Optimization means finding a minima in the high dimension loss surface of the training set in the minimum computational cost and wall-clock time. 

Although optimization advancements sometimes do improve accuracy, we view those as fixing bugs than introducing a feature because, in general, running a low learning rate on a maximally high batch size without any tricks (namely, the entire dataset, also known as Gradient Descent) will eventually find the minimum loss value. Thus, the central challenge of optimization is speed. 

Formally, during training, the goal is to find the best parameters $\theta$ of a neural network. "Best" means that these parameters will yield the minimum loss value, calculated by the loss function $J(\theta)$. The typical loss function for classification, for example, is cross-entropy \citep{goodfellow2016deep}. 

During training, we compute the best direction and relative magnitude, or gradient, along which we should change our parameters $\theta$ by using back-propagation. The gradient is mathematically guaranteed to be the direction of the steepest descent.

\subsection{Optimizers}

Optimizers other than pure GD/SGD, like Momentum, RMSProp, Adam, LARS, and LAMB, include some statistical information about the prior gradients to dynamically adjust the update for faster convergence \citep{tieleman2017divide, kingma2014adam, you2017large}. The goal is to converge more quickly by using this statistical information. 

For one update of parameters $\theta$, optimization algorithms might calculate updates using all or less than all of the training set:

\begin{itemize}
  \item Gradient Descent calculates gradients against the entire dataset and does not analyze statistical information. 
	  \begin{equation*} 
    	\theta = \theta - \eta . \nabla_\theta J(\theta)
	  \end{equation*}
  \item Most optimizers calculate against less than the full dataset, the most basic being Stochastic Gradient Descent (SGD). Note that we consider momentum a differnet optimizer than pure SGD, since momentum includes statistical information about prior gradients. 
	  \begin{equation*} 
	    \theta = \theta - \eta.\nabla_\theta J(\theta; x^{(i:i+n)}; y^{(i:i+n)})
	  \end{equation*}
\end{itemize}

\subsection{Batch Size}

Batch size refers to the number of training examples utilized in one iteration of neural network training. A crucial implementation detail is that the batches must be selected randomly. Unfortunately, Tensorflow's Dataset API’s \href{https://www.tensorflow.org/api_docs/python/tf/data/Dataset}{shuffle} method provides an insufficiently random algorithm

Batch Size is an important hyper-parameter of optimization. The larger the batch size, the more information is contained in each step of back propagation. GPUs are designed to parallelize computations, and the best way to parallelize computation is to increase the batch size. 

Finding the right batch size by trial and error is nearly intractable because it depends on the cardinality and entropy of the dataset, the model architecture, GPU specifications and learning rate. If the batch size is too small, training runs take longer than necessary. And if the batch size is too large, training is using more compute resources than necessary. 

\subsubsection{Intuition about Finding The Right Batch Size}
In Gradient Descent, the gradient used is what we will call true gradient: an average of the the gradients from each and every training example. The true gradient factors in all possible information, by definition. A batch gradient is the average of just a batch of data, rather than the full training data. 

When the batch size is very small, the batch gradient will be noisy estimate of the true network gradient because of high variance in the gradient between batches. Doubling the batch size allows us to train in half the time without additional steps of backprop. By contrast, when the batch size is very large, variance in the gradients between batches is very low. The batch gradient nearly matches the true gradient. Doubling the batch size won't lead to faster training because any two randomly sampled batches will almost have similar gradients.

The optimal batch size, which is a trade off between wall clock time for training and the compute cost in terms of hardware budget, occurs roughly when increasing batch size stops reducing the noisiness of the gradient - where the variance of the gradient is at the same scale as the gradient.

% When the batch size is very small, the batch gradients will have very high variance, and the resulting update will be noisy and stochastic. Doubling the batch size will smooth out the noise, resulting in a Batch Gradient that is a better approximation of the True Gradient. By contrast, when the batch size is very large, the Batch Gradient will almost exactly match the True Gradient, and correspondingly, two randomly sampled batches will have almost the same gradient. As a result, doubling the batch size will barely improve the update – we will use twice as much computation for little gain.

% Transition between the first regime (where increasing the batch size leads to almost perfectly linear speedups) and the second regime (where increasing the batch size mostly wastes computation) should occur roughly where the noise and signal of the gradient are balanced – where the variance of the gradient is at the same scale as the gradient.

\begin{figure}[ht]
\noindent 
\centering{}
\includegraphics[scale=0.2]{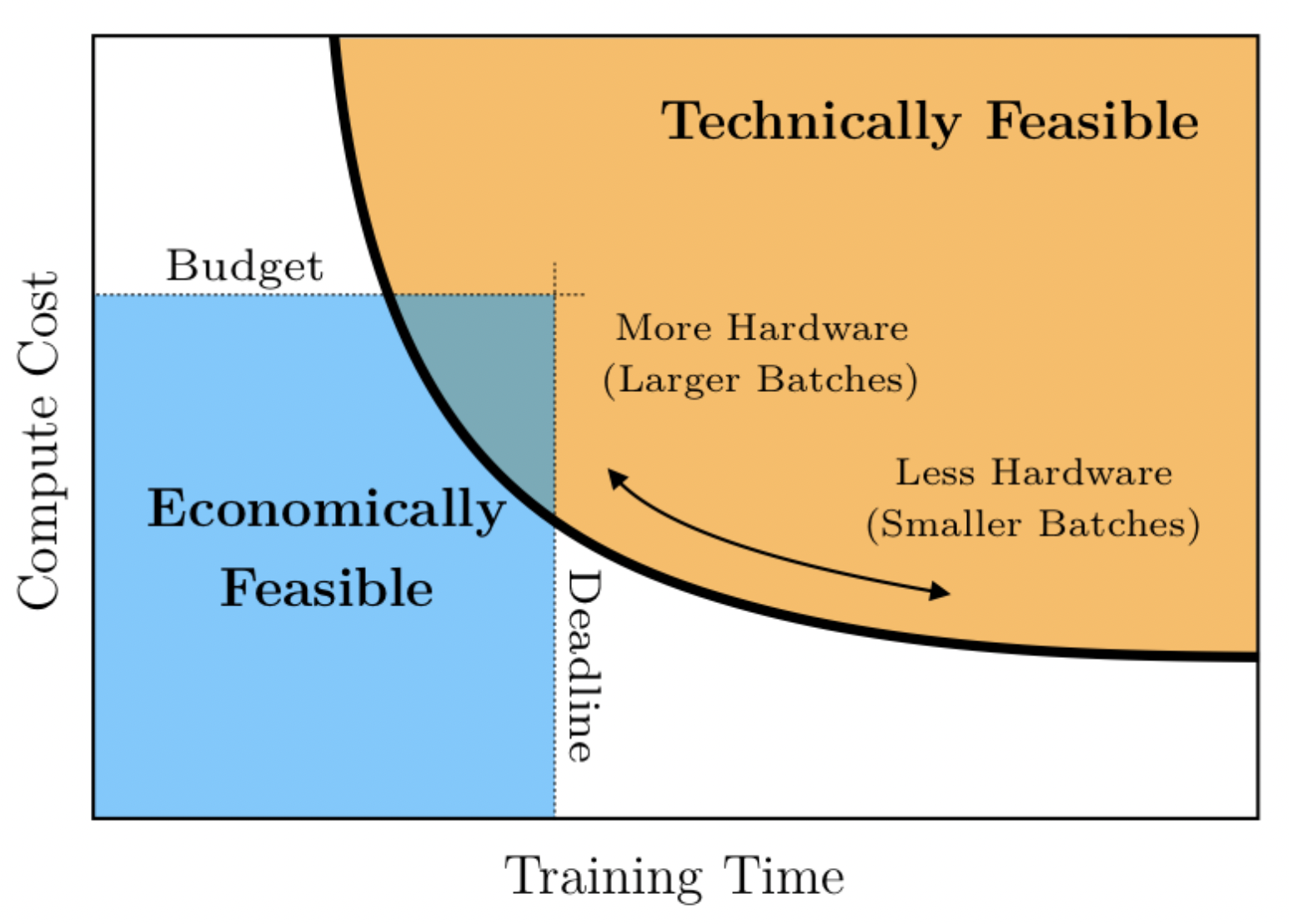}
\caption{Training time and compute cost are primarily determined by the number of optimization steps and the number of training examples processed, respectively (Figure source: ~\citep{mccandlish2018empirical}). \label{fig:trainingtime-computecost}}
\end{figure}

\begin{figure}[ht]
\noindent \centering{}\includegraphics[height=0.25\paperwidth]{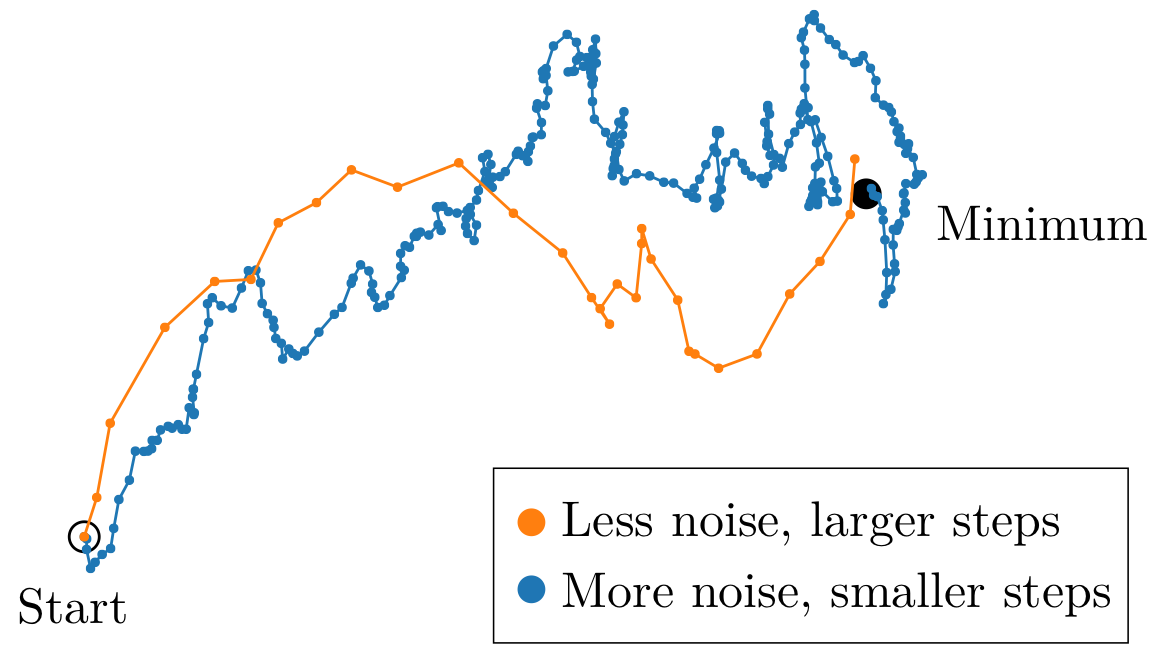}\caption{Comparison between large and small batch training (Figure source: ~\citet{mccandlish2018empirical}). \label{fig:noise-illustration}}
\end{figure}

\subsubsection{The Gradient Noise Scale}
In \citet{mccandlish2018empirical}, the authors distilled the above intuition into a specific algorithm and demonstrated its usefulness across many domains and applications, including MNIST, SVHN, CIFAR10, ImageNet, and Billion Word. It starts by calculating a statistic called the \textbf{gradient noise scale}.

The objective of a deep neural network is to minimize the loss using an optimization algorithm such as stochastic gradient descent. Total loss $L(\theta)$ over the entire distribution is given by an average over loss $L_x(\theta)$ associated with each data point $x$. Therefore $L(\theta) = \mathbb{E}_{x\sim p}[L_x(\theta)]$.

We minimize $L(\theta)$ using SGD-like optimizer by taking repeated steps in the direction of steepest descent of the loss function at the current point. The gradient of the loss function, $G(\theta) = \nabla L(\theta)$, is mathematically guaranteed to be the direction of its steepest descent. Calculating gradient over the entire distribution is expensive. Instead, we obtain an estimate of the true gradient by averaging over a mini-batch of samples of size $B$.

% The loss function is given by an average over a distribution $\rho(x)$ over data point $x$. Each data point $x$ has an associated loss function $L_x(\theta)$ and the full loss is given by $L(\theta) = \mathbb{E}_{x\sim p}[L_x(\theta)]$

% We would like to minimize $L(\theta)$ using an SGD-like optimizer, so the relevant quantity is the gradient $G(\theta) = \nabla L(\theta)$, however calculating the gradient over the entire distribution is time consuming. Instead, we obtain an estimate of the gradient by averaging over a collection of samples from $\rho$

\begin{equation} 
G_{est}(\theta) = \frac{1}{B} \sum_{i=1}^{B}\nabla_\theta L_{x_i}(\theta)\ ; \qquad \qquad
 x_i \sim p
\end{equation}

The above gradient estimate is like a random variable whose expected value (averaged over random batches) is given by the true gradient. Covariance between gradient estimate of different batches is inversely proportional to batch size $B$; As the batch size increases, variance in the gradients between batches is very low.

% The gradient is now a random variable whose expected value (averaged over random batches) is given by the true gradient. Its variance scales inversely with the batch size $B$

\begin{equation} \label{exp_eqn}
\text{E}_{x_1 ... B \sim \rho}[G_{est}(\theta)] = G(\theta)
\end{equation}

\begin{equation} \label{cov_eqn}
\text{cov}_{x_1 ... B \sim \rho}(G_{est}(\theta)) = \frac{1}{B}\Sigma(\theta)
\end{equation}

We want to find the optimal batch size where increasing batch size stops reducing the noisiness of the gradient since noise in the gradient is connected to the maximum improvement in true loss that we can expect from a single gradient update. Let the current gradient update be $V$, it perturbs the model parameters $\theta$ by $\epsilon V$, where $\epsilon$ is the step size. In the next step, we minimize the loss with respect to new model parameters ($\theta - \epsilon V$).

% We are interested in how useful the gradient is for optimization purposes as a function of $B$. We can do this by connecting the noise in the gradient to the maximum improvement in true loss that we can expect from a single gradient update. To start, let $G$ denote the true gradient and $H$ the true Hessian at parameter values $\theta$. If we perturb the parameters $\theta$ by some vector $V$ to $\theta - V$, where $\epsilon$ is the step size, we can expand true loss at this new point to quadratic order in $\epsilon$.

\begin{equation} 
L(\theta - \epsilon V) \approx L(\theta) - \epsilon G^TV + \frac{1}{2} \epsilon^2 V^T H V
\end{equation}

Where $G$ denotes the true gradient and $H$ denotes the true Hessian at parameter values $\theta$.

The above equation is derived from second-order Taylor series approximation of the function $f(x_0 + \delta x)$ is given by $f(x_0 + \delta x) = f(x_0) + \delta x^T + \frac{1}{2} \delta x^T H \delta x$

The step direction is given by a direction proportional to the gradient: $\to \delta x = \alpha g$ ; where $\alpha$ is the learning rate $\Rightarrow f(x_0 + \delta x) \approx f(x_0) + \alpha g^T g + \frac{1}{2} \alpha^{2} g^T H g$. Therefore, minimizing $f(x_0 + \delta x)$ with respect to $\alpha$ we can obtain that this learning rate should be the right term $\alpha = \frac{g^T g}{g^T H g}$. Using same approach, we get $\epsilon_{max} = \frac{|G|^2}{G^T H G}$.

However, in reality we have access only to the noisy estimated gradient $G_{est}$ from a batch of size $B$, thus the best we can do is minimize the expectation $E[L(\theta) - \epsilon G_{est}]$ w.r.t $\epsilon$.

\begin{equation} 
E[L(\theta) - \epsilon G_{est}] = L(\theta) - \epsilon |G|^2 + \frac{1}{2} \epsilon^2 (G^THG + tr(H\Sigma) / B)
\end{equation}

To find the optimal learning rate, minimize the above equation w.r.t $\epsilon$ by setting the partial derivative of the above equation w.r.t $\epsilon$ to 0.

\begin{equation*} 
- |G|^2 + \epsilon (G^THG + \frac{tr(H\Sigma)}{B}) = 0
\end{equation*}

\begin{equation*}
- B|G|^2 + \epsilon (BG^THG + tr(H\Sigma)) = 0
\end{equation*}

\begin{equation*} 
\epsilon = \frac{B|G|^2}{(BG^THG + tr(H\Sigma)} = \frac{|G|^2}{G^THG(1 + \frac{tr(H\Sigma)}{BG^THG}} = \frac{ \frac{|G|^2}{G^THG} }{ (1 + \frac{ \frac{tr(H\Sigma)}{G^THG} }{B}) }
\end{equation*}

\begin{equation} 
\epsilon_{opt}(B) = \frac{ \frac{|G|^2}{G^THG} }{1 + \frac{B_{noise}}{B}}
\end{equation}

Hence, the noise scale is defined as

\begin{equation}
B_{noise} = \frac{tr(H \Sigma)}{G^T H G}
\end{equation}

The Hessian $H$ is a matrix of all possible second derivatives for a function. As the model size increases, the noise scale in the above equation requires some overhead to compute $H$. To simplify noise scale computation, let's assume that the optimization is perfectly well-conditioned i.e, $H$ is a multiple of the identity matrix, then

% The noise scale in the above equation requires some overhead to compute due to the presence of the Hessian $H$. The situation gets even simpler if we make the (unrealistic) assumption that the optimization is perfectly well-conditioned – that the Hessian is a multiple of the identity matrix, then

\begin{equation} \label{bnoise_eqn}
B_{simple} = \frac{tr (\Sigma)}{|G|^2}
\end{equation}

Now the gradient noise scale can be computed with just gradients. Above equation says that the gradient noise scale is equal to the sum of the variances of the individual gradient components, divided by the global norm of the gradient. It is also a statistic to measure how close the estimated gradient is to the true gradient in $L^2$ space. Heuristically, the noise scale measures the variation in the data as seen by the model by taking expectation over individual data points. When the noise scale is small, increasing the batch size becomes redundant, whereas when it is large, we can still learn more from larger batches of data.

It is expensive in terms of wall clock time to calculate the gradient noise mentioned in equation \ref{bnoise_eqn}. In the next section, we discuss how \citet{mccandlish2018empirical} calculate the gradient noise scale with less overhead.

\subsubsection{Unbiased Estimate of the Simple Noise Scale with Less Overhead}
Instead of accumulating gradients of all the batches, which is expensive, we estimate variances of the individual gradient components and global norm of the gradient by accumulating gradients between two batch sizes $B_{small}$ and $B_{big}$ (this could be a multiple of $B_{small}$). Given estimates of $|G_{est}|^2$ for both $B = B_{small}$ and $B = B_{big}$, we can obtain unbiased estimates $|\varrho|^2$ and $S$ for $|G|^2$ and $tr(\Sigma)$ respectively:

Estimate of global norm of the gradient:
\begin{equation} 
|\varrho|^2 \equiv \frac{1}{B_{big} - B_{small}} (B_{big} |G_{B_{big}}|^2 - B_{small} |G_{B_{small}}|^2)
\end{equation}

Estimate of variance of the individual gradient component:
\begin{equation} 
S \equiv \frac{1}{\frac{1}{B_{small}} - \frac{1}{B_{big}}}(|G_{B_{small}}|^2-|G_{B_{big}}|^2)
\end{equation}

The values $S$ and $|\varrho|^2$ computed at every iteration are not unbiased estimators for $tr(\Sigma)$ and $|G^2|$ respectively. To obtain unbiased estimators, we need to calculate the exponentially moving
average $S_{EMA}$ and $|\varrho|^{2}_{EMA}$ as follows

\begin{equation} 
|\varrho|^{2}_{EMA_t} = \alpha |\varrho|^{2}_{t} + (1-\alpha)|\varrho|^{2}_{EMA_{t-1}}
\end{equation}

\begin{equation} 
S_{EMA_t} = \alpha S_t + (1 - \alpha) S_{EMA_{t-1}}
\end{equation}

Where $\alpha $ represents the decay coefficient that weighs the contributions of the currently calculated value and of the previous exponentially moving average. It is an additional hyper-parameter which needs to be tuned. The unbiased estimation of the noise scale, which is a reasonable estimate of the optimal batch size to use during training, is

\begin{equation} 
\hat{B}_{noise} = \frac{S_{EMA}}{|\varrho|^{2}_{EMA}}
\end{equation}

\subsubsection{Results of Gradient Noise Scale}
To validate the results of the paper, we (Masterful) trained EfficientNetB0, using LAMB optimizer, on CIFAR-10 for two different batch sizes. The starting learning rates are different for both batch sizes because a higher batch size requires higher starting learning rate to fully take advantage of potential speedups in the optimization. See \nameref{lr} for a discussion on learning rates. 

Two batch sizes we chose were:
\begin{itemize}
  \item \textbf{64}: A common default value.
  \item \textbf{1024}: This is the batch size predicted by running the above gradient noise scale algorithm.
\end{itemize}

\begin{center}
\begin{tabular}{ | c | c | c | c | c | c | }
  \hline
  batch & learning & training & \# epochs ran within & training & validation\\ 
  size & rate & time & training time & accuracy & accuracy\\ 
  \hline
  64 & 1e-3 & 7416 sec & 82 & 97.92 \% & 70.68 \% \\ \hline
  64 & 7e-3 & 1914 sec & 81 & 98.64 \% & 71.51 \% \\ \hline
\end{tabular}
\end{center}

As we can see from the above table:
\begin{itemize}
  \item The baseline setup achieved 70.68\% accuracy in 82 epochs over 7416 seconds.
  \item The experimental setup, trained with an optimized batch size, achieved equivalent results in 1914 seconds, or \textbf{75\% less wall clock time than the baseline}.
\end{itemize}

\subsection{Learning Rate Policy}
\label{lr}

TBD.

\section{Acknowledgements}

We thank \href{https://scholar.google.com/citations?user=QMkbFp8AAAAJ}{Shibani Santurkar} for her personal correspondence with us regarding her work \citep{santurkar2018does}

We thank Stephen José Hanson for personal correspondence on his discussions with Rumelhart on ideas around weight decay (coauthored with Lorien Pratt) \citep{hanson1988comparing}.

We thank Barrett Zoph for personal correspondence on his paper Resnet-RS (co-authored with Irwan Bello), in which he shared additional details on their ablation study of weight decay in their paper on ResNet-RS. \citep{bello2021revisiting}. 

The format of this report was in part inspired by \citet{papernot2016technical}

\bibliographystyle{unsrtnat}
\bibliography{references}  %%% Uncomment this line and comment out the ``thebibliography'' section below to use the external .bib file (using bibtex) .

\end{document}